\documentclass{article}

\PassOptionsToPackage{numbers}{natbib}
\usepackage[preprint]{neurips_2024}

\usepackage[utf8]{inputenc} 
\usepackage[T1]{fontenc}    
\usepackage{hyperref}       
\usepackage{url}            
\usepackage{booktabs}       
\usepackage{amsfonts}       
\usepackage{nicefrac}       
\usepackage{microtype}      
\usepackage{xcolor}         
\usepackage{graphicx} 
\usepackage{subfig}
\usepackage{pdflscape}

\title{3D-SAR Tomography and Machine Learning for High-Resolution Tree Height Estimation}

\author{%
  Grace Colverd \\
  University of Cambridge \\
  \texttt{gb669@cam.ac.uk}
  \And
   Jumpei Takami \\
  United Nations Office for Outer Space Affairs  \\
  \texttt{jumpei.takami@un.org}
  \And
   Laura Schade \\
UK Department for Energy Security and Net Zero \\
\texttt{laura.schade@energysecurity.gov.uk } \\ 
  \AND
   Karol Bot\\
  INSEC TEC \\
  \And
   Joseph A. Gallego-Mejia\\
Drexel University \\
\texttt{joseph.gallegomejia@drexel.edu}
}

\begin{document}

\maketitle

\begin{abstract}
  
Accurately estimating forest biomass is crucial for global carbon cycle modelling and climate change mitigation. Tree height, a key factor in biomass calculations, can be measured using Synthetic Aperture Radar (SAR) technology. This study applies machine learning to extract forest height data from two SAR products: Single Look Complex (SLC) images and tomographic cubes, in preparation for the ESA Biomass Satellite mission. We use the TomoSense dataset, containing SAR and LiDAR data from Germany's Eifel National Park, to develop and evaluate height estimation models. Our approach includes classical methods, deep learning with a 3D U-Net, and Bayesian-optimized techniques. By testing various SAR frequencies and polarimetries, we establish a baseline for future height and biomass modelling. Best-performing models predict forest height to be within 2.82m mean absolute error for canopies around 30m, advancing our ability to measure global carbon stocks and support climate action.

\end{abstract}

\section{Introduction and Methodology}
We present work modelling forest height from three-dimensional tomographic SAR (TomoSAR) data, developing a robust method for estimating tree heights. The primary objective of this research is to extend our understanding of the benefits of tomography (multiple image acquisition) within SAR research, given the incoming launch of a tomographic SAR satellite (Biomass Satellite mission \citet{esa_esa_nodate}). TomoSAR captures three-dimensional representations of forest structures, requiring multiple SLC images captured from different incidence angles and applied geometric processing including performing a Fourier transformation to create a three-dimensional representation (\citet{ferro-famil_introduction_2017}). This results in a volumetric scattering distribution that provides detailed information about the vertical structure of the forest, beyond traditional SLC images.  TomoSAR datasets are inherently designed for tomographic applications, significantly reducing preprocessing time and complexity, and enabling more immediate and detailed forest structure analysis. We first present details of our models and the different experiments.  We provide further background information and discussion of related works in Appendix \ref{app_intro}.

We use the TomoSense Dataset (\citet{tebaldini_tomosense_2023}) which provides a set of 30 calibrated two-dimensional SLC images from different incidence angles captured by plane, a three-dimensional, processed tomographic cube (tomocube) representing forest scattering and an airborne LiDAR scan for the case study area. The forest canopy sits around 20-35m (see Appendix \ref{app_exp} for 3D visualisation). The 2 SAR bands used in this analysis are L and P (wavelengths of 22cm and 69cm respectively). 2 versions of L-band data are included: monostatic and bistatic acquisition (captured with one plane in a fly-back vs. two planes in tandem). The resolution of L-monostatic, L-bistatic and P bands: (3m, 0.55m, 1.3m), (3m, 0.55m, 2.3m), (5m, 1m, 3m). The LiDAR scan (2D) matches the range and azimuth (x,y) dimensions. The processed tomocube's all have pixel sizes of (321, 665, 36). 

We test the ability of the tomographic cube as input to predict forest height, using LIDAR scans as ground truth. We compare a single-pixel approach with tabular machine learning to an image-based approach using convolution neural networks (CNNs). Within the tabular experiments, we assess spatial autocorrelation effects by comparing the performance of different methods of geographic segmentation of training and test data \citet{salazar_fair_2022}.  We compare performance in height estimation of P-band, L-bistatic and L-monostatic SAR data as well as polarimetries within the bands, comparing single polarimetries (HH, HV, and VV) and union of polarimetries (HH+HV+VV). Further details on the frequency bands and polarization channels are given in Appendix \ref{data}. P-band tomocube covered a marginally larger area than L-band; L-bi and L-mono cover the same area. 

To evaluate metrics with the canopy height measurement taken by the LiDAR, we used the Mean Absolute Error (MAE). MAE measures the average magnitude of errors between predicted and actual canopy heights without considering their direction. This metric is useful when the goal is to minimize the absolute difference in predictions. We also report root mean square error (RMSE) and
the model R-squared (R²) value, another key metric used to assess model performance. An R² value of 1 implies perfect prediction, while a value closer to 0 indicates poor predictive power.

\subsection{Tabular models}

\begin{figure}
    \centering
        \caption{Geographic splits of training/validation/test data. L-r: Swathe, square, quadrant. }
    \label{fig:geosplit}
    \includegraphics[width=0.5\linewidth]{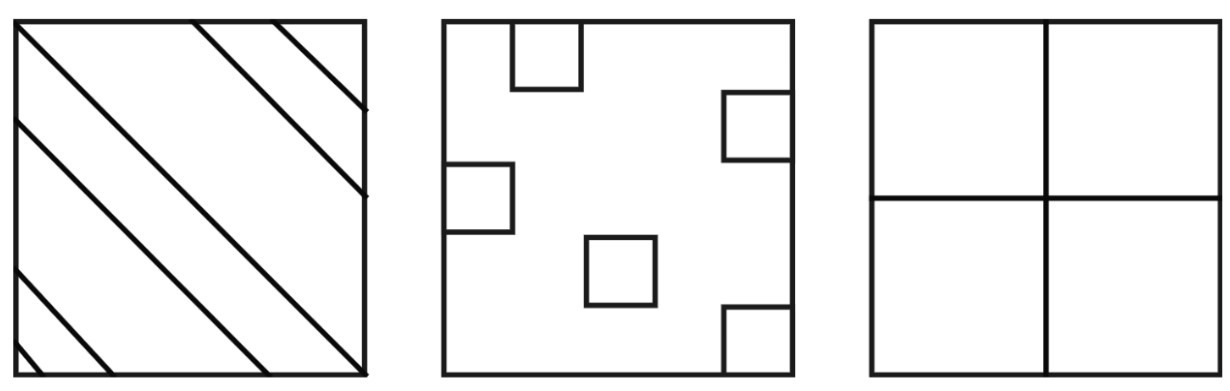}

\end{figure}

Our classical machine learning pipeline uses a (1m x 1m x height) patch as the feature, with LiDAR point clouds as the ground truth. We test three spatial splitting methods for data to quantify how these can impact the model performance. The three methods (swath, square, and quadrant) are shown in Figure \ref{fig:geosplit}. Ratios of training/testing data are 80/20 for square/swathe and 75/25 for quadrant. We flatten the 3D TomoSAR intensity data, converting the cube into tabular data with a separate feature for each height slice, i.e. for a given band $B$ and polarization $ p$, $x,y,z \to x,y, z_1,..,z_n$ where $n= len(z)$. We match the input features with the LiDAR ground truth using a merge on $(x,y)$.  The 1-pixel architecture enables us to use a simple model where each pixel is treated independently and benefits from needing less training data than deep learning approaches. We test including or excluding the azimuth and range coordinates (x,y) as model inputs

For the tabular data we employed AutoGluon (\citet{erickson_autogluon-tabular_2020}), an AutoML tool, to streamline the process of evaluating and selecting the best machine-learning models for tree height estimation for tabular data. AutoGluon simplifies the machine-learning pipeline by automatically tuning hyperparameters, selecting features, and optimizing models for well structured tabular data. AutoGluon uses bayesian optimisation to improve hyperparameter search, and automatically selects validation data. We held out test data until final model selection.

\subsection{CNN models}
Our deep learning pipeline processes the TomoSAR data in its native 3D form and uses a 3D convolution neural network architecture. 
We test three different model architectures adapted for three-dimensional data, all based around the popular U-Net backbones(\citet{ronneberger_u-net_2015}). Broadly they can be characterised as Model 1: Larger U-Net. Model 2: Simple U-Net. Model 3: Simple U-Net with residual connections (inspired by ResNet to improve learning efficiency and model performance (\citet{he_deep_2015}). For each backbone, we tested different methods of collapsing the 3D backbone outputs to a 2D output, including convolution over 36 dimensions, global average pooling along the z dimension, and progressive z reduction through multiple convolutions. The full description of architectures including a diagram is given in Appendix \ref{cnn_arch}.

For dataset generation, we test different patch sizes of $(W,W,z)$ where $W\in \left\{16,32,64\right\}$, $z=36$, and W represents the number of pixels in the range and azimuth directions of a local patch.  Training, validation and test datasets are generated using the quadrant splitting method (with two quadrants allocated to training - see Figure \ref{fig:error_plot}). Quadrant split was chosen to ensure no spatial leakage, and to better represent the challenges the ESA satellite will face. The percentage splits of data are 50/30/20 for train/val/test. All the data is normalised, using a min-max scaler fitted to the training data. Other parameters tested include dropout rates (DR) and batch normalisation. Information on training protocols (implemented with WandB) is in Appendix \ref{model_trainin}. The top-performing model after full parameter selection was based on validation MAE. Test data was held out until the final model was selected. 

\section{Results and Discussion}

\subsection{Geo-Split Comparisons}

The test MAE results for the P-band models is given in Table \ref{tab:comparison_geo}. Including the azimuth and range coordinates (XY) as inputs in the tabular models reduced the MAE across all tabular models, by an average 15\%, 29\% and 49\% across polarisations respectively for square, swath and quadrant geo-splits, emphasizing the risks of spatial autocorrelation.  The increasing impact of geo-location inclusion correlates with the degree of integration between test and training data (refer back to \ref{fig:geosplit}). The difference in performance across the different geographic splits is notable, particularly for square and swathe vs. quadrant. A 47\% drop reduction in quadrant MAE to the average of square and swath MAE to quadrant MAE, and an average change of 35\% change when comparing the relative performance of each of square/swathe/quadrant. The quadrant tabular models overfit the training data and struggle to generalise to the unseen quadrant, even within a relatively homogeneous patch of forest. These findings highlight the critical nature of test data selection and cross-validation in geospatial modelling, with the potential for a 35\% performance swing based solely on test data location choices. 

However, it is prudent to interpret these results cautiously. The performance differences across splits may be artificially inflated due to the small dataset size ((321, 665) pixels), potentially exaggerating the impact of spatial dependencies. These observations underscore the need for rigorous validation strategies in geospatial machine learning, particularly when working with limited datasets. Future work should explore the stability of these findings with larger, more diverse TomoSAR datasets. The superior performance of the CNN model even with less training data suggests that architectures capable of capturing spatial relationships implicitly may offer more robust solutions for this type of geospatial prediction task, and subsequent results all use this pipeline.

\begin{table}
\caption{Test MAE (m) for P-band: Comparison across geographic splits and CNN method. \textit{XY}: Inputs include range \& azimuth. \textit{W/o XY}: Inputs exclude range \& azimuth.}
\label{tab:comparison_geo}
\centering
\begin{tabular}{lccccccc}
\hline
\textbf{Pol Ch.} & \multicolumn{2}{c}{\textbf{Tabular: Square}} & \multicolumn{2}{c}{\textbf{Tabular: Swathe}} & \multicolumn{2}{c}{\textbf{Tabular: Quadrant}} & \textbf{CNN: Quadrant} \\
                 & \textit{XY}   & \textit{W/o XY}   & \textit{XY}   & \textit{W/o XY}   & \textit{XY}     & \textit{W/o XY}    & \textit{W/o XY} \\ \hline
\textbf{HH}      & 3.76          & 4.40              & 4.22          & 4.74              & 4.90            & 7.09               & 2.93            \\
\textbf{HV}      & 2.55          & 2.98              & 2.56          & 3.48              & 4.51            & 6.87               & 2.84            \\
\textbf{VV}      & 2.52          & 3.03              & 2.51          & 3.50              & 4.63            & 6.97               & 2.98            \\ \hline
\end{tabular}
\end{table}

\subsection{Height Estimation}
\begin{figure}[htbp]
    \centering
    \begin{minipage}{0.38\textwidth}
        \centering
         \captionof{table}{Height Estimation for union polarisation channels}
        \label{tab:sar_band_res}

\begin{tabular}{@{}lllll@{}}
\toprule
\textbf{Band}   & \begin{tabular}[c]{@{}l@{}}\textbf{Val} \\ \textbf{MAE} \\ \textbf{(m)}\end{tabular} & \begin{tabular}[c]{@{}l@{}}\textbf{Test} \\ \textbf{MAE}\\ \textbf{(m)}\end{tabular} & \begin{tabular}[c]{@{}l@{}}\textbf{Norm.}\\  \textbf{Test} \\ \textbf{MAE}\end{tabular} & \begin{tabular}[c]{@{}l@{}}\textbf{Test}\\ \textbf{$R^2$}\end{tabular} \\ \midrule
L-Mono & 2.32                                                      & 2.82                                                      & 2.317                                                        & 0.71                                               \\
L-Bi   & 2.31                                                      & 3.07                                                      & 1.33                                                         & 0.68                                               \\
P      & 2.50                                                      & 3.06                                                      & 1.02                                                         & 0.70                                               \\ \bottomrule
\end{tabular}
 
    \end{minipage}%
    \hfill
    \begin{minipage}{0.5\textwidth}
            \caption{Band comparison for MAE normalised by vertical resolution}
        \label{fig:norm_mae}
        \centering
        \includegraphics[width=\linewidth]{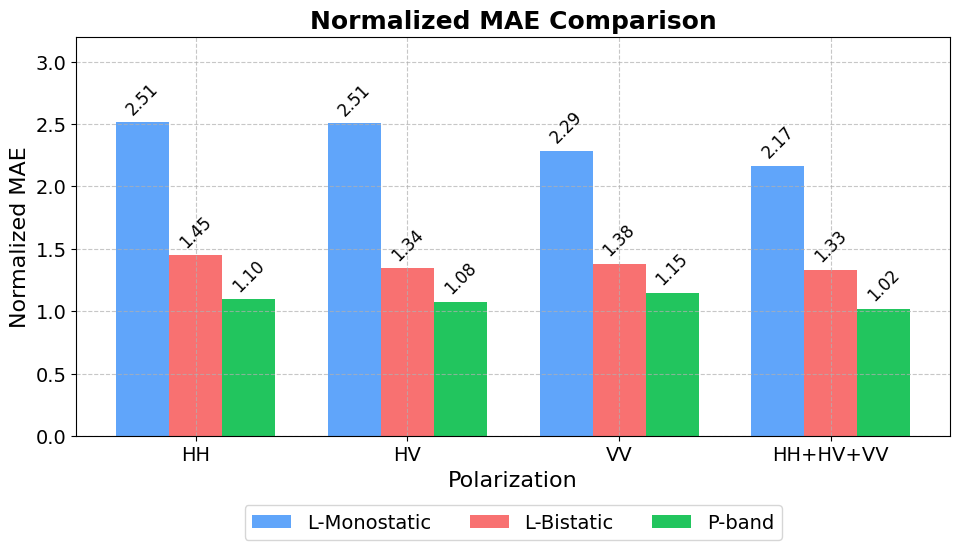}

    \end{minipage}
\end{figure}

The height prediction results for the union of all polarisation channels are given in Table \ref{tab:sar_band_res}. Due to the different vertical resolutions of the bands, we also report Test MAE normalised by the relative vertical resolution to enable cross-band comparisons. The full results including individual polarisations and model settings are given in Appendix \ref{results_appendix}. In absolute terms, L-monostatic achieves the best height predictions with a test MAE of 2.82m (RMSE of 4.21m, model $R^2$ 0.71), marginally better than P and L-bistatic\ . However, after normalising by relative resolutions, the P band has the best performance, with a normalised test MAE of 1.02: a 3.06m MAE with 3m resolution for HH+HV+VV. P band outperforms in relative terms across all polarimetry options, as shown in Figure \ref{fig:norm_mae}. L-bistatic also shows relatively better than L-monostatic, which is to be expected given the greater chance for decoherence in monostatic due to greater time between image acquisition. Within the L band, HV and VV generate better predictions than HH, likely due to stronger returns from vertical objects from the V component. 

Several themes were identified in terms of the successful model architectures and parameters across bands and polarizations. The smaller models 2 and 3 were more successful, likely due to the small amount of training data not enabling full training of the larger model. All of the top-performing models have a patch size of $(16,16,36)$, which seems to offer a good balance between contextual information and maximising the number of samples. All use min-max scaler to standardise the data. Dropout rate ranged from 0 to 20\%. Batch sizes were on the smaller size of those tested, ranging from 2-13, with the majority <10. The collapse method showed a roughly even split between the three tested methods.

\begin{figure}
    \caption{Forest canopy height reconstruction for L-monostatic, HH+HV+VV polarisation. }
    \label{fig:error_plot}
    \centering
    \includegraphics[width=0.95\linewidth]{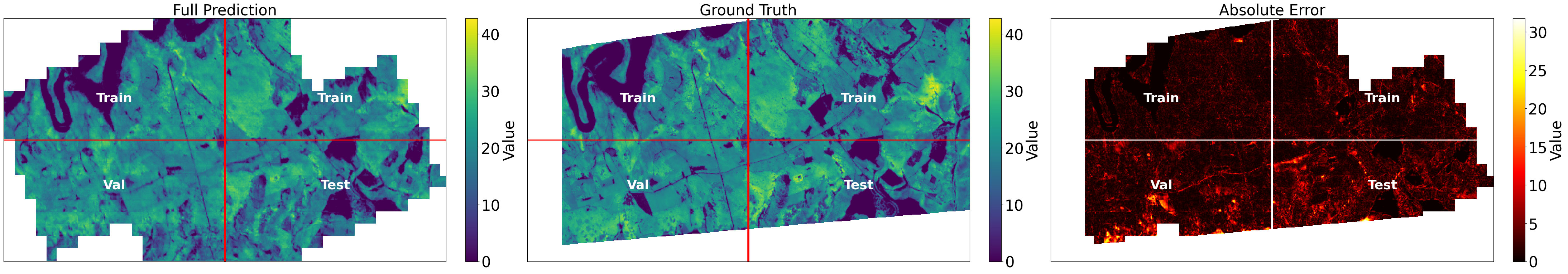}

\end{figure}

Using the L-monostatic HH+HV+VV model, we reconstruct the case study area in Figure \ref{fig:error_plot}. We note the limitations of the current pipeline in dealing with edge effects in patches. Further work should aim to improve this. In addition, the small dataset size limits the ability to robustly test the models transferability: we look forward to testing on other TomoSAR datasets.

\section{Conclusions}
Our study revealed several significant findings. We quantified the impact on model performance of choice of geographic test selection, with an average effect of 35\% on MAE for classical models. This serves as a reminder of the potential for metric manipulation and we suggest the need for robust and transparent data selection procedures, and the use of cross-validation. Geocoding model inputs accounted for an average improvement in performance of 31\% for classical models. The relative performance of the square geo-split in classical modelling suggests the benefit to smaller localised patches of ground truth for improving model performance. 

The customized compact U-Net-based 3D CNN architecture (approximately 1.3M parameters) outperformed classical models in predicting forest height in unknown areas when combined with small patch sizes. Among the different bands and polarisations, L-band demonstrated superior predictive power compared to P-band, due to its greater vertical resolution and reduced sensitivity to ground-level double bounce scattering. When looking at performance relative to vertical resolution, P-band outperformed L-band. 
Our lightweight models offer a significant step forward in the use of Tomographic SAR technology for forestry, offering a scalable and efficient alternative to traditional methods of tree height estimation. The set of models for each band and polarization channels are to be released publicly, and updated as improvements are made.

\begin{ack}
This work has been enabled by FDL Europe | Earth Systems Lab (https://fdleurope.org) a public / private partnership between the European Space Agency (ESA), Trillium Technologies, the University of Oxford and leaders in commercial AI supported by Google Cloud, Scan AI and Nvidia Corporation.

\end{ack}

\bibliographystyle{unsrtnat}
\bibliography{references}


\appendix

\section{Appendix}

\subsection{Scope of the work}\label{app_intro}

Tree height estimation is a critical component of forest management, serving as a key indicator of forest structure, biomass, and overall ecosystem health. Precise tree height data enables forest managers to make informed decisions about timber yield, carbon sequestration potential, and habitat suitability for various species, among other applications. Traditional methods, such as manual measurements and LiDAR surveys, while accurate, often face limitations in cost, time efficiency, and accessibility. The increasing demand for up-to-date and large-scale forest inventory data necessitates more efficient and scalable solutions. Synthetic Aperture Radar (SAR) technology has emerged as a promising alternative, offering the ability to penetrate cloud cover and operate in various weather conditions, making it particularly suitable for monitoring forests in regions with frequent cloud cover, like Northern Europe.

Tomographic SAR (TomoSAR) advances these capabilities by capturing three-dimensional representations of forest structures. TomoSAR requires multiple SLC images captured from different incidence angles and applied geometric processing including performing a Fourier transformation to create a three-dimensional representation. This results in a volumetric scattering distribution that provides detailed information about the vertical structure of the forest, beyond traditional SLC images.

\subsection{Related work}

Estimating tree height using physical models, which often rely on allometric relationships between tree diameter and height, is essential for accurate biomass estimation and ecological modeling (\citet{phalla_importance_2018}). These models, such as the height-diameter relationship model, typically use parameters like diameter at breast height (DBH) to predict tree height (\citet{kershaw_comparison_2017} \citet{de_petris_when_2020}). However, several roadblocks hinder the effectiveness of purely physical models. Data collection challenges arise in dense forests, particularly tropical ones, where closed canopies and the time and cost of obtaining accurate measurements can lead to tree height being overlooked, increasing bias in biomass estimations. Additionally, these models may not adequately account for environmental and spatial variability, which significantly affects tree growth patterns. Traditional methods, like clinometers or hypsometers, introduce measurement errors due to reliance on angle and distance measurements, especially in uneven terrain or when the tree top is obscured (\citet{kershaw_comparison_2017}). Furthermore, while remote sensing technologies like LiDAR provide detailed canopy height models, they often measure the apex rather than the prevailing dendrometric height, leading to discrepancies in actual tree height measurements. These limitations underscore the need for integrating other methods, such as remote sensing and statistical models, to improve the accuracy of tree height estimation (\citet{kearsley_model_2017}).

Satellite and aerial data have become essential tools for estimating tree height in conjunction with machine learning, offering several advantages over traditional ground-based methods. Techniques such as LiDAR, SAR, and optical imagery are commonly used to derive tree height information (\citet{tolan_very_2024}). LiDAR provides high-resolution data by measuring the time it takes for laser pulses to return after hitting the canopy, resulting in accurate digital surface models (DSMs) and digital terrain models (DTMs) ( \citet{ganz_measuring_2019}, 
 \citet{lee_forest_2018}).

Remote sensing methods are advantageous due to their ability to cover large areas quickly and cost-effectively, and they can be enhanced by integrating data from multiple platforms to improve accuracy \citet{lee_forest_2018}, \citet{xuan_intelligent_2023}. However, challenges remain, such as the need for high-quality calibration data and the potential for errors in complex terrains or dense canopies, which can affect the precision of height estimations  (\citet{ganz_measuring_2019}, \citet{lee_forest_2018}). Despite these challenges, advancements in remote sensing technologies continue to improve the reliability and resolution of tree height data, making it a valuable resource for ecological research and forest management.

Commons machine learning methods for predicting tree height include random forests (\citet{wang_modeling_2021}), Gradient Boosting Decision Trees (GBDT) and Support Vector Machines (SVM) have been explored for forest height estimation, with GBDT showing higher precision in certain scenarios (\citet{bao_estimation_2023}) and deep Convolutional Neural Networks (CNNS) (\citet{puliti_tree_2023}). These machine-learning models provide a cost-effective and scalable solution for estimating tree height, although they require high-quality training data and careful calibration to minimize errors.

State-of-the-art works have combined hyperspectral and Lidar data (Sentinel 2 and GEDI) and use probabilistic deep neural networks to generate global tree canopy heights with a balanced RMSE of 7.3m  (\citet{lang_high-resolution_2023}). 
TomoSAR (Tomographic Syn
thetic Aperture Radar) and LiDAR (Light Detection and Ranging) data are increasingly used in conjunction with machine learning models for estimating tree height, offering complementary advantages in capturing vertical forest structure. TomoSAR provides volumetric information by exploiting multiple SAR acquisitions to reconstruct a 3D representation of the forest, which is particularly useful in densely vegetated areas. LiDAR, on the other hand, offers high-resolution point clouds that accurately capture the canopy surface and ground elevation, making it ideal for generating detailed digital elevation models.

Machine learning models such as Random Forest (RF), Gradient Boosting Decision Trees (GBDT), and Support Vector Machines (SVM) have been effectively applied to these datasets. RF is a popular choice due to its robustness and ability to handle large datasets with high dimensionality, making it suitable for integrating TomoSAR and LiDAR data to predict forest height (\citet{wang_modeling_2021}). GBDT, another ensemble learning technique, is used to improve prediction accuracy by sequentially building decision trees that correct errors made by previous models (\citet{bao_estimation_2023}). SVM is employed for its effectiveness in handling nonlinear relationships, which is beneficial when dealing with the complex interactions between TomoSAR and LiDAR features (\citet{bao_estimation_2023}).

Deep learning models, such as YOLOv5, have also been explored, particularly for processing dense LiDAR point clouds. YOLOv5, a convolutional neural network, is used for object detection tasks like identifying tree whorls, which can be indicative of growth patterns and height estimation (\citet{puliti_tree_2023}). These models benefit from transfer learning, allowing them to adapt to new datasets or sensor types with minimal retraining.

AfriSAR and TropiSAR are prominent datasets used for forest structure analysis and biomass estimation. AfriSAR focuses on African tropical forests, while TropiSAR covers tropical forests in French Guiana. Both datasets provide valuable SAR data but are not inherently designed for tomographic applications. In contrast, the TomoSAR dataset, such as the one provided by the upcoming European Space Agency's Biomass Satellite mission, is prebuilt for tomographic uses. This means that TomoSAR data is inherently optimized for generating three-dimensional forest structure information, providing a ready-to-use platform for detailed analysis without the need for extensive preprocessing.

Overall, the integration of TomoSAR and LiDAR data with machine learning models provides a powerful approach for tree height estimation, leveraging the strengths of both data types to improve accuracy and efficiency in forest monitoring and management.

\subsection{Experiment details}\label{app_exp}

In this study, the data is split geographically into training, testing, and validation sets to ensure that the models generalize well to unseen areas. This approach minimizes the likelihood of spatial autocorrelation, which could lead to overly optimistic results if data from similar regions are used across these splits. The geographic split involves segmenting the intensity image and the canopy height map into different regions such as swaths, squares, and quadrants, which are then processed independently.

\subsubsection{Data}\label{data}
The TomoSense SAR bands and frequencies are given in Table \ref{tab:sar_bands}. 

The frequencies within the SAR bands refer to the orientation of the electromagnetic waves transmitted and receive by the emitting device. HH (Horizontal-Horizontal): Both transmitted and received waves are horizontally polarized. 
HV (Horizontal-Vertical): Transmitted wave is horizontal, but received wave is vertical. VV (Vertical-Vertical): Both transmitted and received waves are vertically polarized. This is particularly sensitive to vertical structures and can penetrate vegetation canopies better than HH.

\begin{table}[h]
\caption{TomoSense SAR data \citet{tebaldini_tomosense_2023}}
\label{tab:sar_bands}
\centering
\resizebox{\textwidth}{!}{%
\begin{tabular}{lllllll}
\hline
\textbf{Band} & \textbf{Number} & \textbf{Wavelength} & \textbf{Bandwidth} & \textbf{Slant} & \textbf{Azimuth } & \textbf{Vertical}  \\     
 & \textbf{of passes} & & \textbf{Bandwidth} & \textbf{range resolution} & \textbf{Resolution} & \textbf{ Resolution}
\\ \hline
P    & 28               & 69cm       & 30MHz     & 5m                     & 1m                 & 3m                                                                          \\
L    & 30               & 22cm       & 50MHz     & 3m                     & 55cm               & \begin{tabular}[c]{@{}l@{}}1.3m (monostatic)\\ 2.3m (bistatic)\end{tabular} \\ \hline
\end{tabular}%
}
\end{table}

\begin{figure}
    \caption{3D LiDAR scan of case study area}
    \label{fig:groundtruth_lidar}
    \centering
    \includegraphics[width= 0.8\linewidth]{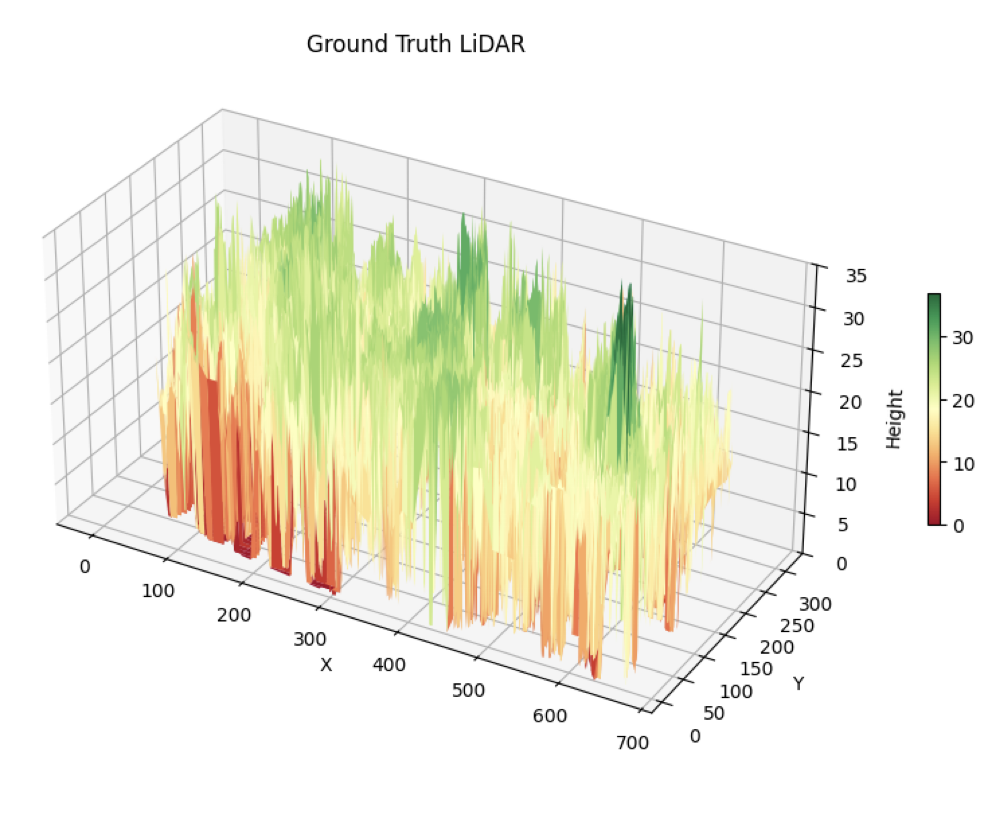}

\end{figure}

\subsubsection{CNN architecture}\label{cnn_arch}

\begin{figure}
    \caption{Model architecture for Models 2 and 3. Block options are given in Figure \ref{fig:encoder-block} and \ref{fig:resid_block}}
    \label{fig:arch}
    \centering
    \includegraphics[width=0.8\linewidth]{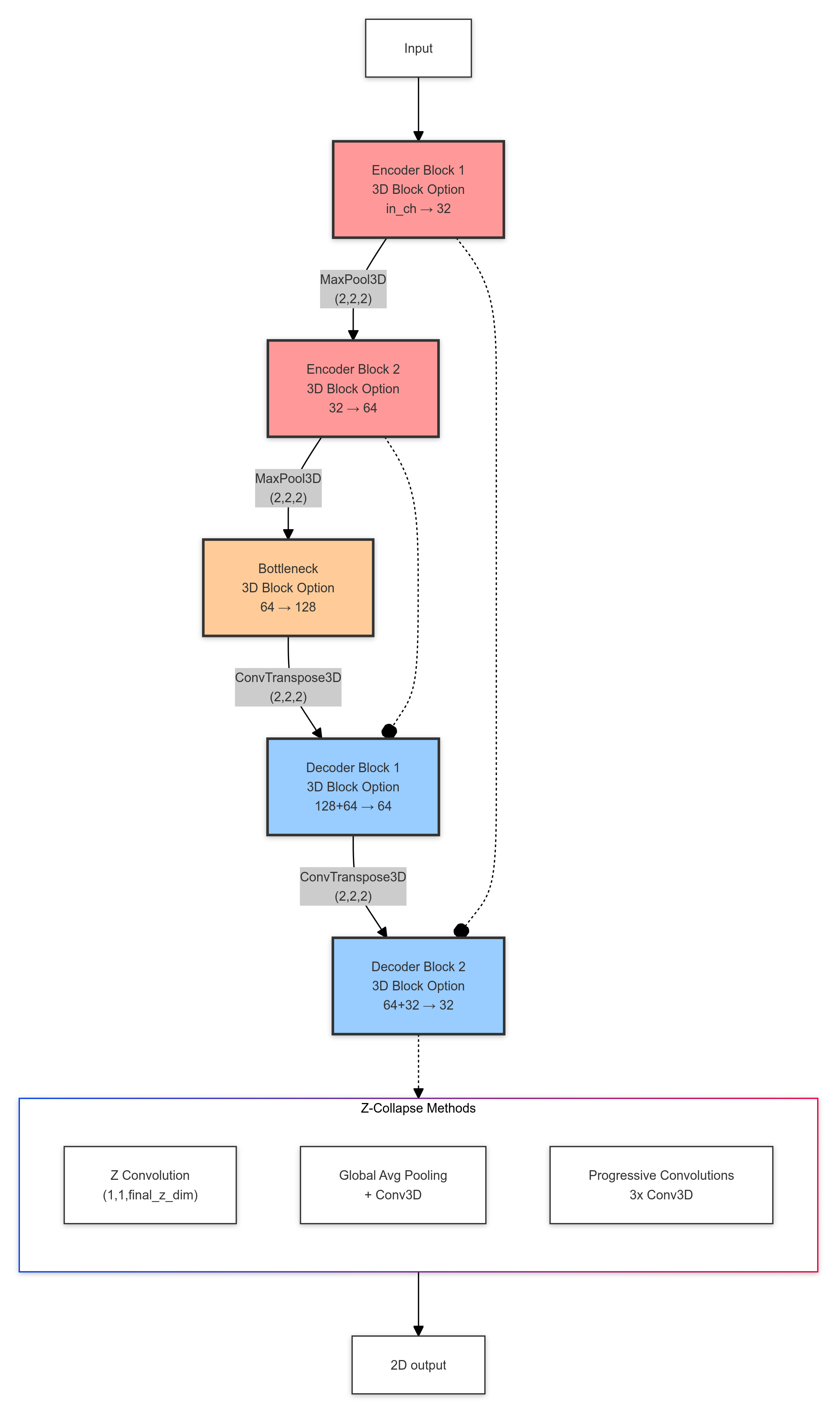}
\end{figure}

Details of the three models tested are given in Table \ref{tab:model-arch}. The diagram for the architecture of the two successful models is given in Figure \ref{fig:arch}. 

\begin{table}[h]
\caption{Model Architectures}
\label{tab:model-arch}
\centering
\begin{tabular}{lll}
\hline
\textbf{Model Name} & \textbf{Backbone} & \textbf{Parameter Count} \\ \hline
Model 1             & U-Net, 3 encoder/decoder layers         & 21M  \\
Model 2             & U-Net, 2 encoder/decoder layers         & 1.3M \\
Model 3             & U-Net, 2 encoder/decoder layers with residual connections & 1.2M \\ \hline
\end{tabular}
\end{table}

\begin{figure}[htbp]
    \centering
    \includegraphics[width=0.2\textwidth]{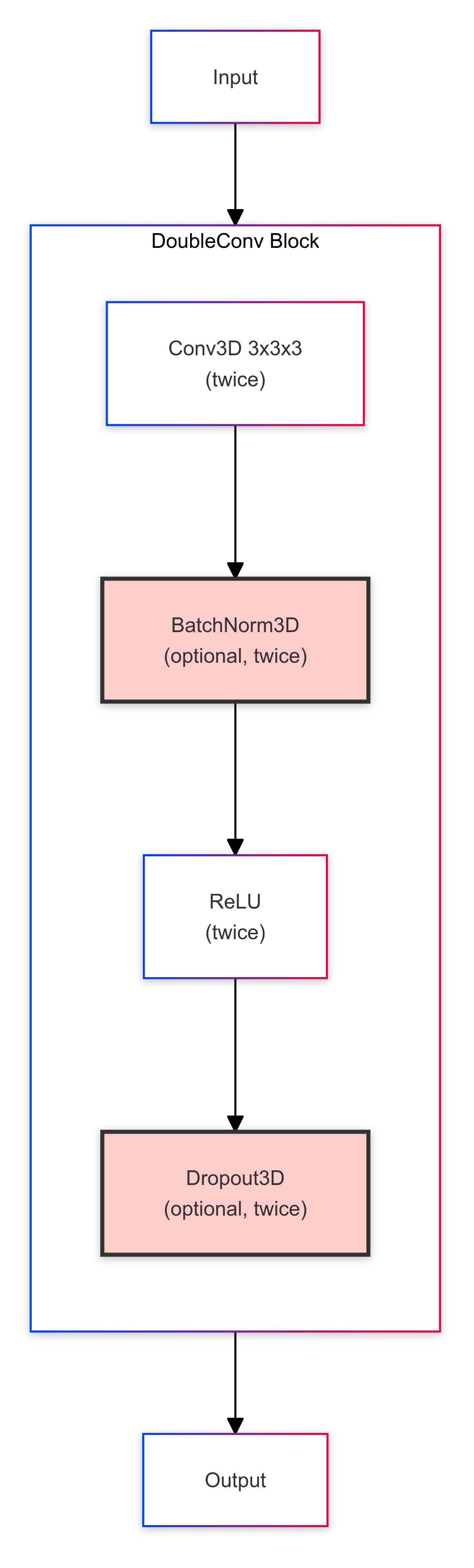}
    \caption{Double Convolution block used in Model 2. Batch Norm optional. Dropout: rate varied from 0-0.5.}
    \label{fig:encoder-block}
\end{figure}

\begin{figure}[htbp]
    \centering
    \includegraphics[width=0.35\textwidth]{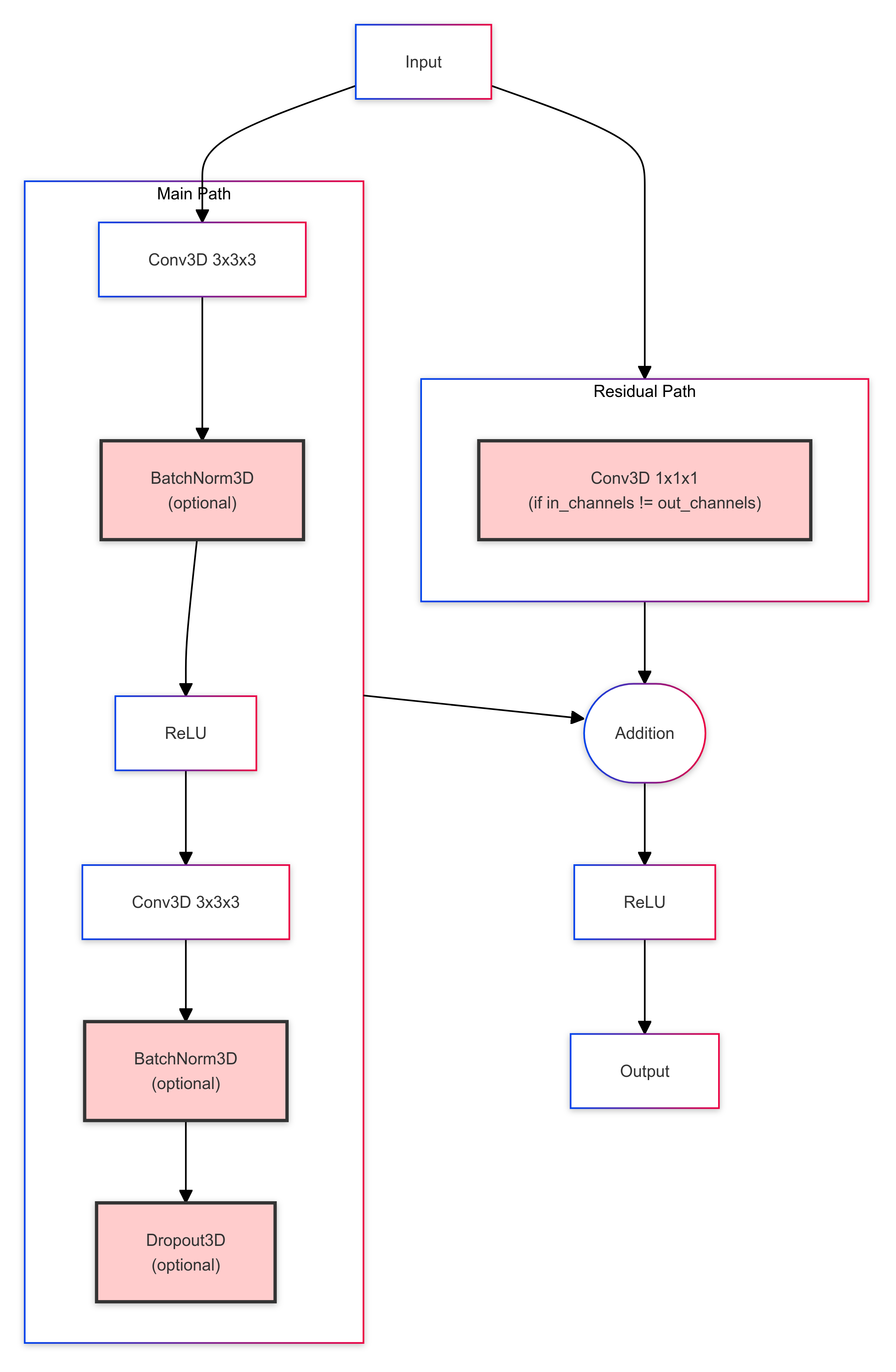}
    \caption{Residual block used in Model 3. Batch Norm optional. Dropout: rate varied from 0-0.5.}
    \label{fig:resid_block}
\end{figure}

\subsubsection{Training software and hardware}\label{model_trainin}

All models are coded in PyTorch. We test different architecture configurations using the parameters mentioned (model backbone, collapse method, patch size, polarization) for each band as well as traditional hyperparameter tuning on batch size (BS), number of epochs and learning rate (LR). Early stopping based on validation MAE was implemented. Sweeps are run using WandB, with Bayesian search. All models are trained using MSE as the loss function, with ADAM optimiser. Models are trained on Google Cloud using a Tesla GPU T4 with 16gb memory. Due to the small size of the dataset, training was quick on the order of 5-30 minutes and we ran extensive hyperparameter sweeps, with at least 300 runs for each band/polarisation option.

\subsection{Additional results }\label{results_appendix}

The full model parameters for all band options and polarisation are given in Table \ref{tab:cnn_results}. 

\begin{landscape}
\begin{table}[]
\caption{Best performing height estimation CNN models. All settings use patch size (16,16,36), batch normalisation and min-max scaler.}
\label{tab:cnn_results}
                                                 
\begin{tabular}{llllllllllll}
\hline
Band         & Pol      & Model & LR          & Z-Collapse                                                  & BS & DR  & Best & Val & Val & Test& Test \\ 
         &       &  &           &  & BS & DR  &  Epoch &  MAE &  RMSE &  MAE &  RMSE \\ \hline

L-Bistatic   & HH       & 2     & 0.000097135 & Global Av Pool                                              & 7  & 0   & 61         & 2.45    & 3.66     & 3.24     & 4.67      \\
L-Bistatic   & HV       & 2     & 0.000070058 & Conv                                                        & 7  & 0.1 & 56         & 2.37    & 3.67     & 3.09     & 4.54      \\
L-Bistatic   & VV       & 3     & 0.000099036 & \begin{tabular}[c]{@{}l@{}}Progresive\\  Conv\end{tabular}  & 9  & 0.2 & 75         & 2.31    & 3.57     & 3.17     & 4.66      \\
L-Bistatic   & Union    & 2     & 0.000082879 & Conv                                                        & 11 & 0.1 & 60         & 2.31    & 3.54     & 3.07     & 4.50      \\
L-Monostatic & HH       & 3     & 0.000098292 & Conv                                                        & 13 & 0.2 & 65         & 2.75    & 4.16     & 3.27     & 4.67      \\
L-Monostatic & HV       & 2     & 0.000074736 & Conv                                                        & 9  & 0.1 & 31         & 2.72    & 4.17     & 3.26     & 4.76      \\
L-Monostatic & VV       & 2     & 0.000091559 & \begin{tabular}[c]{@{}l@{}}Progresive\\  Conv\end{tabular}  & 5  & 0   & 33         & 2.59    & 4.08     & 2.97     & 4.40      \\
L-Monostatic & Combined & 2     & 0.00004583  & \begin{tabular}[c]{@{}l@{}}Progressive\\  Conv\end{tabular} & 8  & 0.2 & 108        & 2.32    & 3.80     & 2.82     & 4.21      \\
P            & HH       & 3     & 0.000098284 & \begin{tabular}[c]{@{}l@{}}Global Av\\  Pool\end{tabular}   & 3  & 0.2 & 59         & 2.93    & 4.23     & 3.29     & 4.72      \\
P            & HV       & 2     & 0.000083692 & Conv                                                        & 7  & 0.1 & 40         & 2.84    & 4.22     & 3.23     & 4.72      \\
P            & VV       & 2     & 0.000083689 & \begin{tabular}[c]{@{}l@{}}Progressive\\  Conv\end{tabular} & 2  & 0.2 & 52         & 2.98    & 4.53     & 3.44     & 4.96      \\
P            & Union    & 2     & 0.000048659 & \begin{tabular}[c]{@{}l@{}}Progressive\\  Conv\end{tabular} & 2  & 0.2 & 61         & 2.50    & 3.92     & 3.06     & 4.56      \\ \hline
\end{tabular}
\end{table}
\end{landscape}

\begin{figure}[htbp]
        \caption{Comparison of ground truth and height reconstruction (VV-Monostatic) }
        \label{fig:groundtruth}
    \centering
    \begin{minipage}{0.48\textwidth}
        \centering
        \includegraphics[width=\linewidth]{figs/grountruth_hegiht.png}

    \end{minipage}%
    \hfill
    \begin{minipage}{0.48\textwidth}
        \centering
        \includegraphics[width=\linewidth]{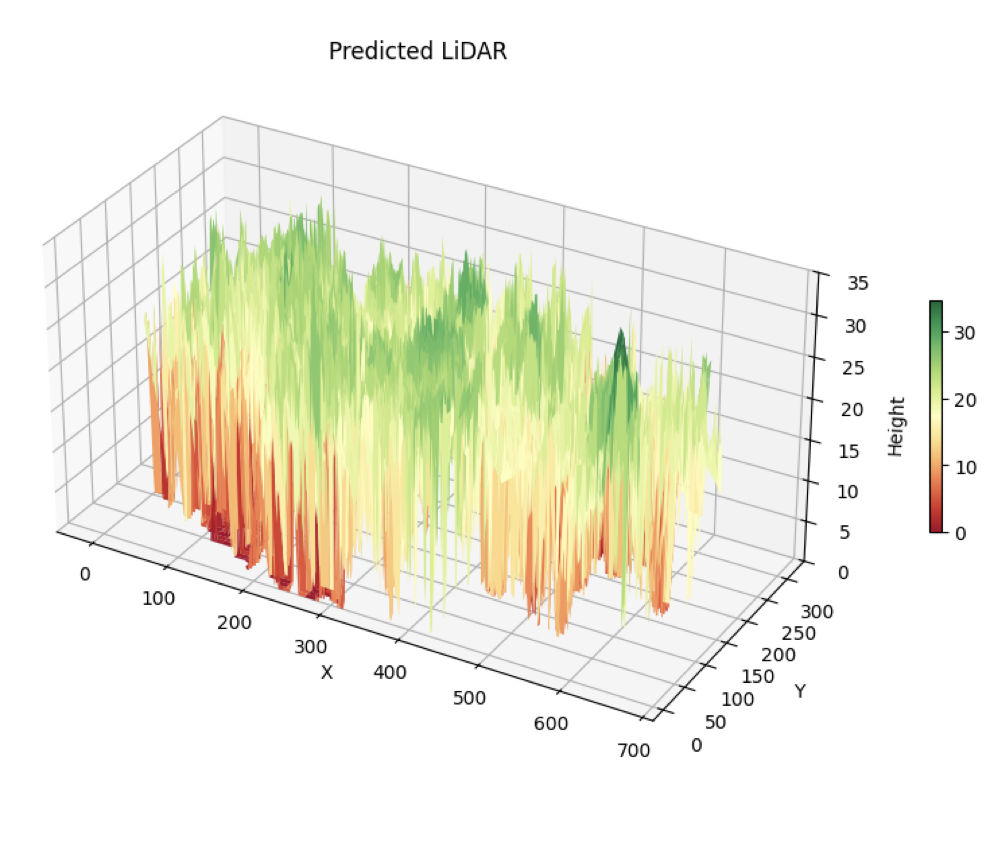}

    \end{minipage}
\end{figure}

\end{document}